\documentclass{article}

\usepackage{arxiv}
\usepackage{fullpage}
\usepackage[utf8]{inputenc}
\usepackage[T1]{fontenc}
\usepackage[inline]{enumitem}
\usepackage{hyperref}
\usepackage{url}
\usepackage{amsfonts}
\usepackage{clrscode}
\usepackage{physics}
\usepackage{graphicx}
\usepackage{caption}
\usepackage{float}
\usepackage{subcaption}
\usepackage{siunitx}
\usepackage{mathtools}
\usepackage{comment}
\usepackage[title]{appendix}

\usepackage{amsmath,amsfonts,amssymb,amsthm}

\usepackage{subcaption}
\usepackage{siunitx}
\usepackage{booktabs}
\usepackage{etoolbox}
\usepackage{graphicx}
\usepackage[table,xcdraw]{xcolor}
\usepackage{comment}
\usepackage{amsmath,amsfonts,amssymb,amsthm}
\usepackage{nicefrac}
\usepackage{booktabs}

\renewrobustcmd{\bfseries}{\fontseries{b}\selectfont}
\renewrobustcmd{\boldmath}{}
\newrobustcmd{\B}{\bfseries}

\usepackage{acro}
\DeclareAcronym{dbtai}{
short = d-BTAI ,
long  = dynamic-Binary Tree Anomaly Identifier ,
}
\DeclareAcronym{pef}{
short = PEF ,
long  = Parametric Elliot Function ,
}
\DeclareAcronym{ef}{
short = EF ,
long  = Elliot Function ,
}
\DeclareAcronym{wtwme}{
 short = WTM ,
 long  = Weighted Time-Window Moving Estimation ,
 }
 \DeclareAcronym{cblof}{
short = CBLOF ,
long  = Cluster-Based Local
Outlier Factor  ,
}
\DeclareAcronym{lc}{
short = LC ,
long  = Large Cluster ,
}
\DeclareAcronym{sc}{
short = SC ,
long  = Small Cluster ,
}
 \DeclareAcronym{ecblof}{
short = ECBLOF ,
long  = Enhanced Cluster Based Local Outlier Factor ,
}
 \DeclareAcronym{mri}{
short = MRI ,
long  = Magnetic Resonance Imaging ,
}
\DeclareAcronym{fn}{
short = FN ,
long  = False Negative ,
}
\DeclareAcronym{uvmgbtai}{
short = UVMGTree ,
long  = Uni-variate Multi-Generations Tree,
}
\DeclareAcronym{scsd}{
short = SCSD ,
long  = Small Cluster Single Data Instance ,
}

\DeclareAcronym{pso}{
short = PSO ,
long  = Particle Swarm Optimization ,
}
\DeclareAcronym{lstm}{
short = LSTM ,
long  = Long Short Term Memory networks ,
}
\DeclareAcronym{fp}{
short = FP ,
long  = False Positive ,
}
\DeclareAcronym{tp}{
short = TP ,
long  = True Positive ,
}
\DeclareAcronym{svm}{
short = SVM ,
long  = Support Vector Machine ,
}
\DeclareAcronym{ocsvm}{
short = OCSVM ,
long  = One Class Support Vector Machine ,
}
\DeclareAcronym{bat}{
short = BAT ,
long  = Anomaly Tree ,
}
\DeclareAcronym{bata}{
short = BAT ,
long  = Binary Anomaly Tree ,
}
\DeclareAcronym{if}{
short = iForest ,
long  = Isolation Forest ,
}
\DeclareAcronym{auc}{
short = AUC ,
long  = Area under the ROC curve  ,
}
\DeclareAcronym{rwf}{
short = RWF ,
long  = Rolling Window Forecasting ,
}
\DeclareAcronym{ewf}{
short = EWF ,
long  = Expanding Window Forecasting ,
}
\DeclareAcronym{roc}{
short = ROC ,
long  = Receiver Operating Characteristics ,
}

\DeclareAcronym{ppv}{
short = PPV ,
long  = Positive Predictive Value ,
}
\DeclareAcronym{npv}{
short = NPV ,
long  = Negative Predictive Value ,
}
\DeclareAcronym{sla}{
short = SLA ,
long  = Service Level Agreement ,
}
\DeclareAcronym{dl}{
short = DL ,
long  = Deep Learning,
}
\DeclareAcronym{sp}{
 short = SP ,
 long  = Specificity ,
 }
\DeclareAcronym{mgbtai}{
 short = MGTree ,
 long  = Multi-Generations Tree,
 } 
 \DeclareAcronym{ewma}{
 short = EWMA ,
 long  = Exponentially Weighted Moving Average,
 }
 \DeclareAcronym{wma}{
 short = WMA ,
 long  = Weighted Moving Average,
 }
 \DeclareAcronym{tn}{
short = TN ,
long  = True Negative ,
}
\DeclareAcronym{cnn}{
short = CNN ,
long  = Convolutional  Neural Network ,
}
\DeclareAcronym{msma}{
short = MSMA ,
long  = Multi-Stage Memetic Algorithm ,
}
\DeclareAcronym{at}{
short = AT ,
long  = Anomaly Tree ,
}
\DeclareAcronym{msmbtai}{
short = MSMBTAI ,
long  = Multi-Stage Memetic Binary Tree Anomaly Identifier ,
}
\DeclareAcronym{iforest}{
short = iForest ,
long  = Isolation Forest ,
}
\DeclareAcronym{lof}{
short = LOF ,
long  = Local Outlier Factor ,
}
\DeclareAcronym{msmvmca}{
short = MSMVMCA ,
long  = Multi-Stage Multi-Version Memetic Clustering Algorithm ,
}
 \DeclareAcronym{wtwai}{
 short = WTWAI ,
 long  = Weighted Time-Window Anomaly Identifier ,
 }

\hypersetup{
    colorlinks=true,
    linkcolor=blue,
    filecolor=magenta,      
    urlcolor=cyan,
}

\title{quantile-LSTM: A Robust LSTM for Anomaly Detection in Time Series Data}

\author{
   Snehanshu Saha \\
  Department of CSIS and APPCAIR \\
  Birla Institute of Technology and Science\\
  Goa, India \\
  \texttt{snehanshus@goa.bits-pilani.ac.in}\\
   \And
  Jyotirmoy Sarkar \\
 Department of CSIS  \\
  Birla Institute of Technology and Science\\
  Goa, India \\
  \texttt{jyotirmoy208@gmail.com}\\
   \And
    Soma Dhavala \\
  Founder\\
  MLSquare\\
  Bengaluru, India \\
  \texttt{soma@mlsquare.org} \\
    \And
  Santonu Sarkar \\
  Department of CSIS \\
  Birla Institute of Technology and Science\\
  Goa, India \\
  \texttt{santonus@gmail.com}\\
   \And
 Preyank~Mota \\
  Department of CSIS \\
  Birla Institute of Technology and Science\\
  Goa, India \\
  \texttt{f20190331@goa.bits-pilani.ac.in}\\
}

\begin{document}
\date{}
\maketitle

\begin{abstract}
Anomalies refer to the departure of systems and devices from their normal behaviour in standard operating conditions. An anomaly in an industrial device can indicate an upcoming failure, often in the temporal direction. In this paper, we make two contributions: 1) we estimate conditional quantiles and consider three different ways to define anomalies based on the estimated quantiles. 2) we use a new learnable activation function in the popular \ac{lstm}  architecture to model temporal long-range dependency. In particular, we propose \ac{pef} as an activation function (AF) inside LSTM, which saturates lately compared to \emph{sigmoid} and \emph{tanh}.  The proposed algorithms are compared with other well-known anomaly detection algorithms, such as \ac{if}, Elliptic Envelope, Autoencoder, and modern Deep Learning models such as Deep Autoencoding Gaussian Mixture Model (DAGMM), Generative Adversarial Networks (GAN). The algorithms are evaluated in terms of various performance metrics, such as Precision and Recall. The algorithms have been tested on multiple industrial time-series datasets such  as  Yahoo,  AWS, GE,  and machine sensors. We have found that the LSTM-based quantile algorithms are very effective and outperformed the existing algorithms in identifying anomalies.  

\end{abstract}

\section{Introduction}\label{sec:introduction}

Anomalies indicate a departure of a system from its normal behaviour. In Industrial systems, they often lead to failures. By definition, anomalies are rare events. As a result, from a Machine Learning standpoint, collecting and classifying anomalies pose significant challenges. For example, when anomaly detection is posed as a classification problem, it leads to extreme class imbalance (data paucity problem). Though several current approaches use semi-supervised neural network to detect anomalies~\cite{Forero:2019,Sperl2020}, these approaches 
%
still require some labeled data. In the recent past, there have been approaches that attempt to model normal dataset and consider any deviation from the normal as an anomaly. For instance, autoencoder-based family of models ~\cite{Jinwon:2015} use some form of thresholds to detect anomalies. Another class of approaches relied on reconstruction errors~\cite{Sakurada:2019}, as an anomaly score. If the reconstruction error of a datapoint is higher than a threshold, then the datapoint is declared as an anomaly. However, the threshold value can be specific to the domain and the model, and deciding the threshold on the reconstruction error can be cumbersome. 

In this paper, we have introduced the notion of {\em quantiles} in multiple versions of the LSTM-based anomaly detector. Our proposed approach is principled on:\\
\begin{itemize}
    \item training models on a normal dataset
    \item modeling temporal  dependency
    \item proposing an adaptive solution that does not require manual tuning of the activation
\end{itemize}

Since our proposed model tries to capture the normal behavior of an industrial device, it does not require any expensive dataset labeling. Our approach also does not require re-tuning of threshold values across multiple domains and datasets. We have exhibited through empirical results later in the paper (see Table \ref{table:datasetschar} of Appendix \ref{appendix:datasetchar} 
) that the distributional variance does not impact the prediction quality.
Our contributions are three folds: \\
\textbf{(1)} Introduction of {\em quantiles}, free from the assumptions on data distributions, in design of quantile-based LSTM techniques and their application in anomaly identification.\\
\textbf{(2)} Proposal of the {\em Parameterized Elliot} as a 'flexible-form, adaptive, learnable' activation function in LSTM, where the parameter is learned from the dataset. Therefore, it does not require any manual retuning when the nature of the dataset changes. We have shown empirically that the modified LSTM architecture with \ac{pef} performed better than the \ac{ef} and showed that such behavior might be attributed to the slower saturation rate of \ac{pef}.\\
\textbf{(3)} Demonstration of {\em superior performance} of the proposed \ac{lstm} methods over state-of-the-art (SoTA) deep learning (Autoencoder ~\cite{Yin}, DAGMM~\cite{zong2018deep}, DevNet~\cite{pang2019deep}) and non-deep learning algorithms (\ac{if}~\cite{iforest}, Elliptic envelope ~\cite{envelope})

The rest of the paper is organized as follows.  The proposal and discussion of various LSTM-based algorithms are presented in section \ref{sec:varlstm}. Section \ref{sec:background} describes the LSTM structure and introduces the \ac{pef}. This section also explains the intuition behind choosing a parameterized version of the AF and better variability due to it. Experimental results are presented in section \ref{sec:experiment}. Section \ref{related} discusses relevant literature in anomaly detection. We conclude the paper in section \ref{sec:conclusion}.

\section{Anomaly detection with Quantile LSTMs}\label{sec:varlstm}
Since {\em distribution independent} and {\em domain independent} anomaly detection are the two key motivation behind this work, we borrow the concept of quantiles from Descriptive and Inferential Statistics to address this challenge.
\subsection{Why Quantile based approach?}
Quantiles are used as a robust alternative to classical conditional means in Econometrics and Statistics ~\cite{koenker}. In a previous work, Tambwekar et.al.[~\cite{Tambwekar:2022} extended the notion of conditional quantiles to the binary classification setting, allowing to quantify the uncertainty in the predictions and provide interpretations into the functions learnt by the models via a new loss called binary quantile regression loss (sBQC). The estimated quantiles are leveraged to obtain individualized confidence scores that provide an accurate measure of a prediction being misclassified. Since quantiles are a natural choice to quantify uncertainty, they are a natural candidate for anomaly detection. However, to the best of our knowledge, quantile based method has not been used for anomaly detection, however natural it seems.
\par Empirically, if the data being analyzed are not actually distributed according to an assumed distribution, or if there are other potential sources for anomalies that are far removed from the mean, then quantiles may be more useful descriptive statistics than means and other moment-related statistics. Quantiles can be used to identify probabilities of the range of normal data instances such that data lying outside the defined range are conveniently identified as anomalies.

The important aspect of distribution-free anomaly detection is the anomaly threshold being agnostic to the data from different domains. 
Simply stated, once a threshold is set (in our case, 10-90), we don't need to tune the threshold in order to detect anomalous instances for different data sets. Quantiles allows using distributions for many practical purposes, including looking for confidence intervals. Quantile divides a probability distribution into areas of equal probability i.e. quantiles offer us to quantify chances that a given parameter is inside a specified range of values. This allows us to determine the confidence level of an event (anomaly) actually occurring. 

Though the mean of a distribution is a useful measure when it is symmetric, there is no guarantee that actual data distributions are symmetric. If there are potential sources for anomalies are far removed from the mean, then medians are more robust than means, particularly in skewed  and heavy-tailed data. It is well known that quantiles minimize check loss~\cite{Horowitz:1992}, which is a generalized version of Mean Absolute Error (MAE) arising from medians rather than means. Thus, quantiles have less susceptibility to long-tailed distributions and outliers, in comparison to mean~\cite{Dunning2021}. 


Therefore, it makes practical sense to investigate the power of quantiles in detecting anomalies in data distributions. Unlike the methods for anomaly detection in the literature, our proposed quantile-based thresholds applied in the quantile-LSTM are generic and not specific to the domain or dataset. The need to isolate anomalies from the underlying distribution is significant since it allows us to detect anomalies irrespective of the assumptions on the underlying data distribution. We have introduced the notion of quantiles in multiple versions of the LSTM-based anomaly detector in this paper, namely (i) quantile-LSTM (ii) iqr-LSTM
and (iii) Median-LSTM. 
All the LSTM versions are based on estimating the quantiles instead of the mean behaviour of an industrial device. Note, the median is $50\%$ quantile. 

\subsection{Various quantile-LSTM Algorithms}
\begin{figure*}[!htp]
  \centering
    \includegraphics[width=0.8\textwidth]{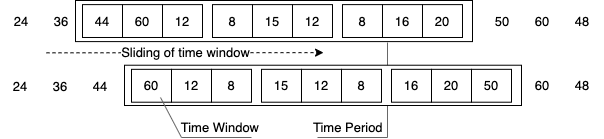}
\caption{Sliding movement of a time period}
\label{fig:movement}
\end{figure*}

Before we discuss quantile-based anomaly detection, we describe the data structure and processing setup, with some notations. Let us consider $x_i, i=1,2,..,n$ be the $n$ time-series training datapoints. We consider $T_k = \{ x_i: i=k,\cdots,k+t \}$ be the set of $t$ datapoints, and let $T_k$ be split into $w$ disjoint windows with each window of integer size $m=\frac{t}{w}$ and  $T_k = \{ T_k^1,\cdots,T_k^w \}$. Here, $T_k^j = \{ x_{k+m (j-1)},...,x_{k+ m(j)-1}\}$. In Figure~\ref{fig:movement}, we show the sliding characteristics of the proposed algorithm on a hypothetical dataset, with $t=9,m=3$. Let $Q_{\tau}(D)$ be the sample quantile of the datapoints in the set $D$. The training data consists of, for every $T_k$, $X_{k,\tau} \equiv \{Q_{\tau}(T_k^j)\}, j=1,\cdots,w$ as predictors with $y_{k,\tau} \equiv Q_{\tau}(T_{k+1})$, sample quantile at a future time-step, as the label or response. Let $\hat{y}_{k,\tau}$ be the predicted value by an LSTM model.

A general recipe we are proposing to detect anomalies is to: (i) estimate quantile $Q_{\tau}(x_{k+t+1})$ with $\tau \in (0,1)$ and (ii) define a statistic that measures the outlier-ness of the data, given the observation $x_{k+t+1}$. Instead of using global thresholds, thresholds are adaptive i.e. they change at every time-point depending on quantiles.

\subsubsection{quantile-LSTM} As the name suggests, in quantile-LSTM, we forecast two quantiles $q_{low}$ and $q_{high}$  to detect the anomalies present in a dataset. We assume the next quantile values of the time period after sliding the time period by one position are dependent on the quantile values of the current time period. 
\begin{figure*}[!htb]
\centering
\begin{subfigure}{\textwidth}
  \centering
  \includegraphics[width=0.85\linewidth]{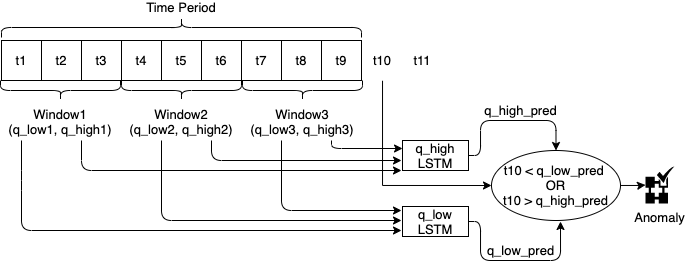}  
  \caption{Anomaly detection process using quantile-LSTM}
  \label{fig:quantilelstm}
\end{subfigure}
\begin{subfigure}{\textwidth}
  \centering
  \includegraphics[width=0.85\linewidth]{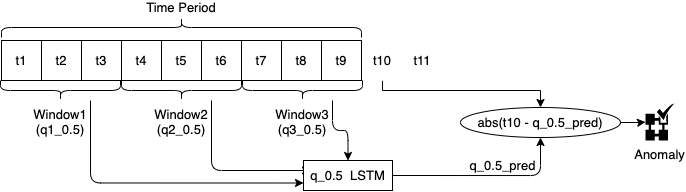}  
  \caption{Anomaly detection process using median-LSTM}
  \label{fig:mediumlstm}
\end{subfigure}
\caption{ Sigmoid function has been applied as an recurrent function, which is applied on the outcome of the forget gate ($f_t=\sigma(W_f*[h_{t-1},x_t]+b_f)$) as well as input gate ($ i_t=\sigma(W_i*[h_{t-1},x_t]+b_i)$). \ac{pef} decides the information to store in cell $ c\hat{} _{t}=PEF(W_c*[h_{t-1},x_t]+b_c)$.}
\end{figure*}

 It is further expected that, nominal range of the data can be gleaned from $q_{low}$ and $q_{high}$. Using these $q_{low}$ and $q_{high}$ values of the current time windows, we can forecast $q_{low}$ and $q_{high}$ values of the next time period after sliding by one position. Here, it is required to build two LSTM models, one for $q_{low}$ (LSTM$_{qlow}$) and another for $q_{high}$ (LSTM$_{qhigh}$). Let's take the hypothetical dataset as a training set from Figure \ref{fig:quantilelstm}. It has three time windows from time period  $x_1\cdots x_9$. Table \ref{table:firsttp} defines the three time windows of the time period $x_1\cdots x_9$ and the corresponding $q_{low}$, $q_{high}$ values against the time window.
\begin{table}[!htpb]
 \centering
 \begin{tabular}{c c c }
\hline
TW &$q_{low}$ & $q_{high}$ \\
\hline
$x_1,x_2,x_3$  & $X_{1,low} \equiv Q_{low}(T_1^1)$ & $X_{1,high} \equiv  Q_{high}(T_1^1)$\\
\hline
$x_4,x_5,x_6$  & $X_{2,low} \equiv Q_{low}(T_1^2)$ & $X_{2,high} \equiv  Q_{high}(T_1^2)$\\
\hline
$x_7,x_8,x_9$  & $X_{3,low} \equiv Q_{low}(T_1^3)$ & $X_{3,high} \equiv  Q_{high}(T_1^3)$\\
\hline
\end{tabular}
\caption{The first time period and its corresponding time windows}
\label{table:firsttp}
\end{table}

The size of the inputs to the LSTM depends on the number of time windows $w$  and one output. Since three time windows have been considered for a time period in this example, both the LSTM models will have three inputs and one output. For example, the LSTM predicting the lower quantile, would have $X_{1,low}$, $X_{2,low}$, $X_{3,low}$ as its puts and $y_{1,low}$ as its output, for one time-period. A total of $n-t+1$ instances will be available for training the LSTM models assuming no missing values.

After building the LSTM models, for each time period it predicts the corresponding quantile value and slides one position to the next time period on the test dataset. quantile-LSTM applies a following anomaly identification approach. If the observed value $x_{k+t+1}$ falls outside of the predicted $(q_{low}, q_{high)}$, then the observation will be declared as an anomaly. For example, the observed value $x_{10}$ will be detected as an anomaly if $x_{10}< \hat{y}_{1,low}$ or $x_{10} > \hat{y}_{1,high}$. Figure \ref{fig:quantilelstm} illustrates the anomaly identification technique of the quantile-LSTM on a hypothetical test dataset. 
\par
\subsubsection{IQR-LSTM} iqr-LSTM is a special case of quantile-LSTM where $q_{low}$ is 0.25 and $q_{high}$ is the 0.75 quantile. In addition, another LSTM model predicts median $q_{0.5}$ as well. Effectively, at every time index $k$, three predictions are made $\hat{y}_{k,0.25},\hat{y}_{k,0.5}, \hat{y}_{k,0.75}$. Based on this, we define the Inter Quartile Range (IQR) $\hat{y}_{k,0.75} - \hat{y}_{k,0.25}$. Using IQR, the following rule identifies an anomaly when
$x_{t+k+1} > \hat{y}_{k,0.5} + \alpha (\hat{y}_{k,0.75} - \hat{y}_{k,0.25})$ or 
$x_{t+k+1} < \hat{y}_{k,0.5} - \alpha (\hat{y}_{k,0.75} - \hat{y}_{k,0.25})$

\subsubsection{Median-LSTM} Median-LSTM, unlike quantile-LSTM, does not identify the range of the normal datapoints; rather, based on a single LSTM, distance between the observed value and predicted median ($x_{t+k+1}-\hat{y}_{k,0.5}$) is computed, as depicted in Figure \ref{fig:mediumlstm}, and running statistics are computed on this derived data stream. The training set preparation is similar to quantile-LSTM. 

To detect the anomalies, Median-LSTM uses an implicit adaptive threshold. It is not reasonable to have a single threshold value for the entire time series dataset when dataset exhibits seasonality and trends. We introduce some notations to make description concrete. Adopting the same conventions introduced before, define $d_k \equiv x_{t+k+1}-Q_{0.5}(T_{k+1}), k=1,2,\hdots,n-t$ and partition the difference series into $s$ sets of size  $t$ each, i.e., $D \equiv {D_p, p=1,\hdots,s}$, where $D_p = \{ d_i: i=(s-1)t+1,\hdots,st \}$. After computing the differences on the entire dataset,  for every window $D_p$, mean ($\mu_p$) and standard deviation ($\sigma_p$) for the individual time period $D_p$. As a result, $\mu_p$ and $\sigma_p$ will differ from one time period to another time period. Median-LSTM detects the anomalies using upper threshold and lower threshold parameters of a particular time period $D_p$ and they are computed as follows: 
$$T_{p,lower}=\mu_p+w\sigma_p; T_{p,higher}=\mu_p-w\sigma_p$$
An anomaly can be flagged for $d_k \in T_p$ when either $d_k > T_{p,higher}$ or $d_k < T_{p,lower}$  Now, what should be the probable value for $w$? If we consider $w=2$, it means that any datapoint beyond two standard deviations away from the mean on either side will be considered as an anomaly. It is based on the intuition that differences of the normal datapoints should be close to the mean value, whereas the anomalous differences will be far from the mean value. 
Hence 95.45\% datapoints are within two standard deviations distance from the mean value. It is imperative to consider $w=2$ since there is a higher probability of the anomalies falling into the 4.55\% datapoints. We can consider $w=3$ too where 99.7\% datapoints are within three standard deviations. However, it may miss the border anomalies, which are relatively close to the normal datapoints and only can detect the prominent anomalies. Therefore we have used $w=2$ across the experiments.

\subsection{Probability Bound}
 In this subsection, we analyze different datasets by computing the probability of occurrence of anomalies using the quantile approach. We have considered 0.1, 0.25, 0.75, 0.9, and 0.95 quantiles and computed the probability of anomalies beyond these values, as shown in Table \ref{table:probabilitybound}
  of appendix section.  
 The multivariate datasets are not considered since every feature may follow a different quantile threshold. Hence it is not possible to derive a single quantile threshold for all the features. It is evident from Table  \ref{table:probabilitybound} of Appendix \ref{appendix:pobabilitybound} of 
 that the probability of a datapoint being an anomaly is high if the datapoint's quantile value is either higher than 0.9 or lower than 0.1. However, if we increase the threshold to 0.95, the probability becomes 0 across the datasets. This emphasizes that a higher quantile threshold does not detect anomalies. It is required to identify the appropriate threshold value, and it is apparent from the table that most of the anomalies are nearby 0.9 and 0.1 quantile values. Table \ref{table:probabilitybound} 
 also demonstrates the different nature of the anomalies present in the datasets. For instance, the anomalies of Yahoo Dataset$_1$ to Yahoo Dataset$_6$ are present nearby the quantile value 0.9, whereas the anomalies in Yahoo Dataset$_7$ to Yahoo Dataset$_9$ are close to both quantile values 0.9 and 0.1. Therefore, it is possible to detect anomalies by two extreme quantile values. We can consider these extreme quantile values as higher and lower quantile thresholds and derive a lemma. We provide a proof in the appendix section.

\textbf{Lemma 1:} For an univariate dataset $\mathcal{D}$, 
the probability of an anomaly $\mathcal{P(A)}=\mathcal{P}(\mathcal{E} > \alpha_{high}) +\mathcal{P(F}<\alpha_{low})$, where $\alpha_{high}, \alpha_{low}$ are the higher and lower level quantile thresholds respectively.

The lemma entails the fact that anomalies are trapped outside the high and low quantile threshold values. The bound is independent of data distribution as quantiles assume nominal distributional characteristics. 

\section{LSTM with Parameterized Elliot Activation (\ac{pef})}\label{sec:background}
We introduce the novel parameterized Elliot activation function 
 \ac{pef}, an adaptive variant of usual activation, wherein we modify the LSTM architecture by replacing the activation function of the LSTM gates with \ac{pef} as follows.

 A single LSTM block is composed of four major components: an input gate, a forget gate, an output gate, and a cell state. We have applied the parameterized Elliot Function (PEF) as activation. 
 \subsection{Parameterized Elliot Function \ac{pef}}
 \ac{pef} is represented by  
\begin{equation}\label{eq:pef}
   f(x)= \frac{\alpha x}{1+|x|}
\end{equation}
with the first order derivative of \ac{pef} as: $f'(x)=\frac{\alpha}{(|x|+1)^2}$. The function is equal to 0, and the derivative is also equal to the parameter $\alpha$ at the origin. After the introduction of the PEF, the hidden state equation is:$h_t=O_t\alpha_c PEF(C_t)$. By chain rule, $$\frac{\partial J}{\partial \alpha_c}=\frac{\partial J}{\partial \alpha_c}=\frac{\partial J}{\partial h_t}O_t *Elliot(C_t)$$. After each iteration, the $\alpha_c$ is updated by gradient descent $\alpha_c^{(n+1)}=\alpha_c^n+\delta*\frac{\partial J}{\partial \alpha_c}$ (See appendix \ref{appendix:backpropa} for back propagation of \ac{lstm} with \ac{pef}). 
Salient features of the PEF are:
\begin{enumerate}
    \item The $\alpha$ in equation \ref{eq:pef} is learned during the back-propagation like other weight parameters of the LSTM model. Hence, this parameter, which controls the shape of the activation, is learned from data. Thus, if the dataset changes, so does the final form of the activation, which saves the ``parameter tuning'' effort.
%
%
    \item The cost of saturation of standard activation functions impedes training and prediction, which is an important barrier to overcome. While PEF derivative also saturates as the $|x|$ increases, the saturation rate is less than other activation functions, such as $\tanh$, $sigmoid$.
    \item \ac{pef} further decreases the rate of saturation in comparison to the non-parameterized Elliot function.
\end{enumerate}
To the best of our knowledge, insights on ’learning’ the parameters of an activation function are not available in literature except for the standard smoothness or saturation properties activation functions are supposed to possess. It is, therefore, worthwhile to investigate the possibilities of learning an activation function within a framework or architecture that uses the inherent patterns and variances from data. 
\begin{figure*}[!htb]
\centering
\begin{subfigure}{.45\textwidth}
 \includegraphics[width=\textwidth]{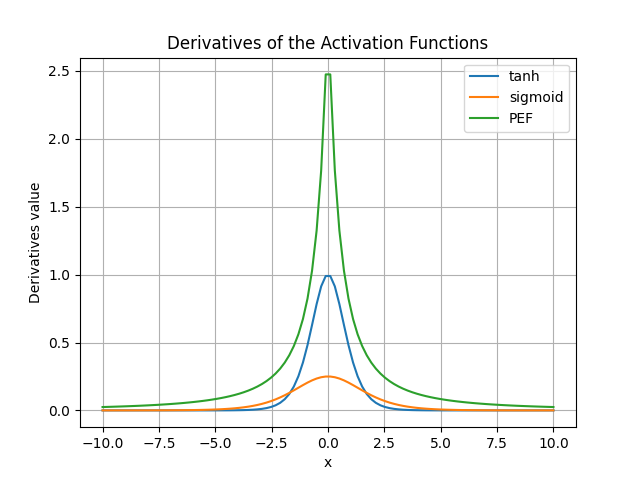}
\caption{Derivatives comparisons of various activation functions. }
\label{fig:activationcomp}
\end{subfigure}
\begin{subfigure}{.45\textwidth}
  \includegraphics[width=\textwidth]{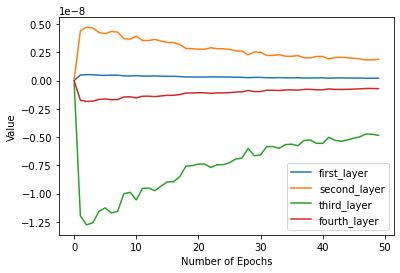}
\caption{LSTM values for 4 layers and 50 epochs using PEF as activation function using AWS2.}
\label{fig:pefplot}
\end{subfigure}
\begin{subfigure}{.45\textwidth}
   \includegraphics[width=\textwidth]{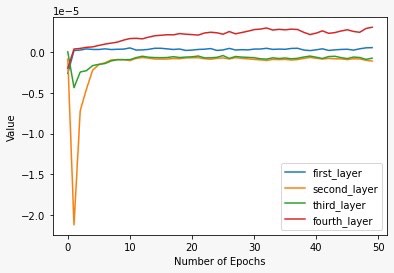}
\caption{LSTM values for 4 layers and 50 epochs using Sigmoid as activation function using AWS2.}
\label{fig:sigmoidplot}
  \end{subfigure}
\begin{subfigure}{.45\textwidth}
   \includegraphics[width=\textwidth]{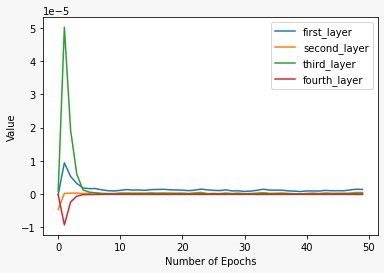}
\caption{LSTM values for 4 layers and 50 epochs using Tanh as activation function using AWS2.}
\label{fig:tanhplot}
\end{subfigure}
\begin{subfigure}{.45\textwidth}
  \includegraphics[width=\linewidth]{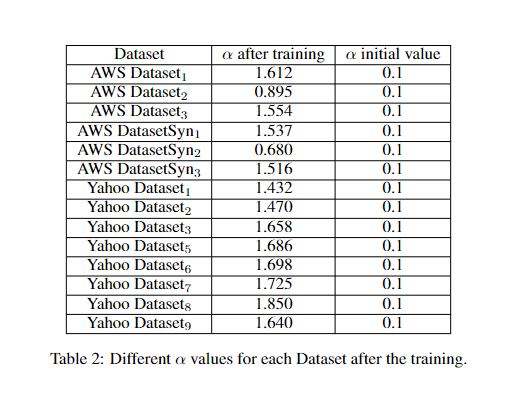}  
  \label{fig:sub-first}
\end{subfigure}
\begin{subfigure}{.45\textwidth}
  \includegraphics[width=\textwidth]{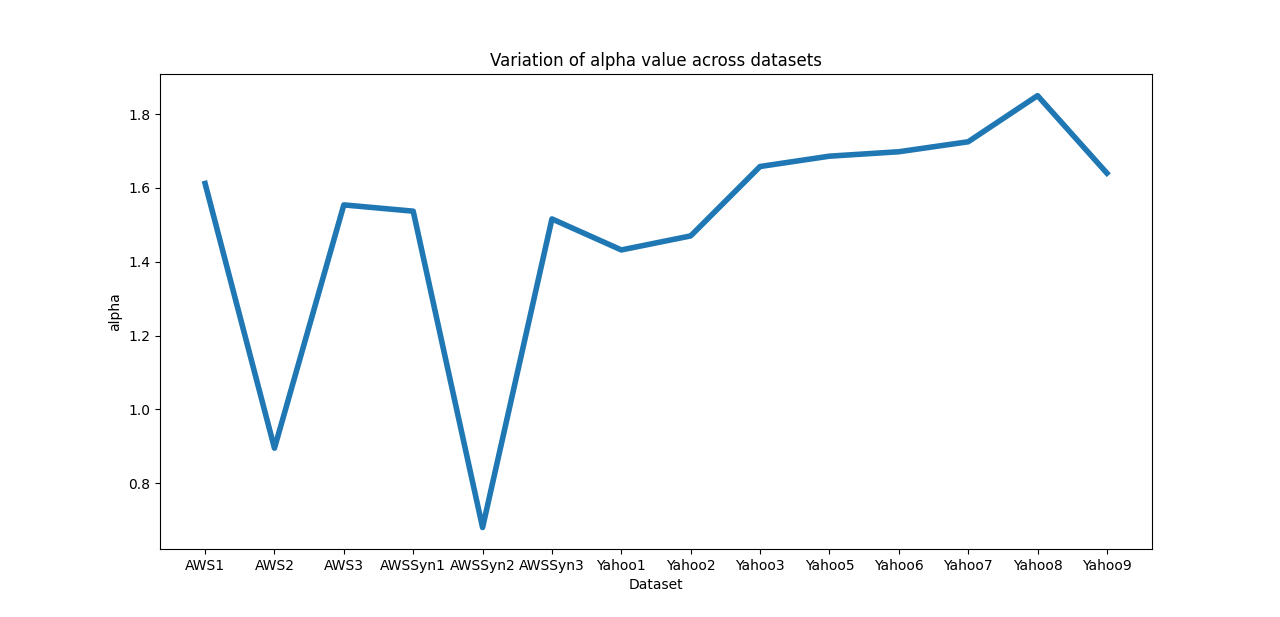}
\caption{The final $\alpha$ values learn on each dataset. We can see the final $\alpha$ value is different for different datasets.}
\label{fig:betaplot}
\end{subfigure}

\caption{Slow saturation rate as well as behavioral comparison of the different layers of \ac{lstm} model after the introduction of \ac{pef} with other activation functions. It also shows the final value of the learned parameter $\alpha$ on various datasets.
}
\end{figure*}
\setlength\belowcaptionskip{-1ex}
\subsection{PEF saturation} The derivative of the PEF is represented by: $    =\frac{\alpha}{x^2}EF^2$. While the derivatives of the sigmoid and tanh are dependent on x, PEF is dependent on both $\alpha$ and x. Even if $\frac{EF^2(x)}{x^2}$ saturates, the learned parameter $\alpha$ will help the \ac{pef} escape saturation. The derivatives of the sigmoid, tanh saturate when $x>5$ or $x<-5$. However, it is not true with PEF as evident from fig \ref{fig:activationcomp}.
As empirical evidence, the layer values for every epoch of the model are captured using various activation functions like \ac{pef}, sigmoid and tanh. It is observed that, after about 10 epochs, the values of the layers becomes more or less constant for sigmoid and tanh (fig \ref{fig:sigmoidplot} and fig \ref{fig:tanhplot}), indicating their values have already saturated whereas for PEF, variation can be seen till it reaches 50 epochs (fig  \ref{fig:pefplot}). This shows that in comparison to sigmoid and tanh as activation functions, PEF escapes saturation due to its learned parameter $\alpha$. \textit{The parameter $\alpha$ in \ac{pef}} changes its value as the model trains over the training dataset while using PEF as the activation function. Since it is a self training parameter, it returns different values for different datasets at the end of training. These values have been documented in table 2 and plotted in fig \ref{fig:betaplot}.  
Table 2 demonstrates the variations in $\alpha$  values across multiple datasets as these values get updated. 

\section{Experiment}\label{sec:experiment}
In this section, we have evaluated the performance of the quantile-LSTM techniques on multiple datasets. We have identified multiple baseline methods, such as \ac{if}, Elliptic envelope, Autoencoder and several deep learning based approaches for comparison purposes (See section \ref{related} for more details on baseline methods). 
\footnote{\ac{lstm} code: https://github.com/PreyankM/Quantile-LSTM}
\subsection{Datasets}
The dataset properties have been shown in Table \ref{table:datasetschar} of Appendix \ref{appendix:datasetchar}. A total of 29 datasets, including real industrial datasets and synthetic datasets, have been considered in the experiments. The industrial datasets include Yahoo webscope \footnote{https://webscope.sandbox.yahoo.com/}, AWS cloudwatch \footnote{https://github.com/numenta/NAB/tree/master/data}, GE. There are a couple of datasets with either one or few anomalies, such as AWS$_1$, AWS$_2$. We have injected anomalies in AWS, Yahoo, and GE datasets to produce synthetic data for fair comparison purposes. The datasets are univariate, unimodal or binodal and follow mostly Weibull, Gamma and Log normal distribution. The highest anomaly percentage is 1.47 (GE Dataset$_2$), whereas AWS Dataset$_2$ has reported the lowest percentage of anomaly i.e. 0.08 (For more details see Table \ref{table:datasetschar} 
2 of Appendix \ref{appendix:datasetchar}
). 


\subsection{Results-Industrial Datasets}
Table \ref{table:quantilecomp} demonstrates the performance comparison of various LSTM techniques. Precision and Recall, two performance metrics, are shown in the table. The Median-LSTM has achieved Recall 1 in most datasets (10 out of 15 datasets). In comparison to existing benchmarks, LSTM methods are SOTA on most of the datasets in terms of Recall. For comparison purposes, we have first compared the Recall. If the Recall is the same for two different methods, then we have compared the Precision. The method which has a higher Recall and Precision will be considered as a better performer. In AWS datasets, most of the techniques have achieved the highest Recall apart from DAGMM and DevNet. DevNet needs minimum two anomalies hence it is not applicable for AWS1 and AWS2. However, as per Precision, iqr-LSTM has performed better than other methods. In the case of GE$1$, DevNet has produced a better result, whereas quantile based LSTM techniques has outperformed others on GE$_2$. Median-LSTM has demonstrated better result in Ambient temperature. In the case of Yahoo datasets, Median-LSTM has achieved the highest Recall on four datasets; however, quantile-LSTM and iqr-LSTM have produced better results on several datasets. For example, Median-LSTM and iqr-LSTM both achieved Recall 1 on Yahoo$_1$. However, if we compare the Precision, iqr-LSTM has shown better results. It is evident from the table \ref{table:quantilecomp} that all these LSTM versions are performing very well on these industrial datasets. We compared our method with a recent anomaly detection method based on Graph Neural Network (GNN)~\cite{deng2021graph} 
We observe that GNN has not shown superior performance in comparison to the quantile based technique. For example, GNN's recall value  is less in comparison to the recall value of 1 quantile based techniques have produced (on AWS2, AWS3, Yahoo1, Yahoo2, Yahoo9). In terms of precision, GNN produced better results than quantile LSTM only on two datasets, namely, Yahoo1 and Yahoo9. 
\begin{table*}[!htb]
\renewcommand\thetable{3}
 \centering
 \resizebox{\textwidth}{!}{
 \begin{tabular}{|c|c|c|c|c|c|c|c|c|c|c|c|c|c|c|c|c|c|c|c|}
\hline
    Dataset&Anomaly&  \multicolumn{2}{c|}{iqr-LSTM}&\multicolumn{2}{c|}{Median-LSTM}&\multicolumn{2}{c|}{quantile-LSTM}&\multicolumn{2}{c|}{Autoencoder}&\multicolumn{2}{c|}{GAN} &\multicolumn{2}{c|}{DAGMM} &\multicolumn{2}{c|}{DevNet} &\multicolumn{2}{c|}{\ac{if}} &\multicolumn{2}{c|}{Envelope}    \\
   \hline
      &  & Precision &Recall& Precision &Recall&Precision &Recall&Precision &Recall&Precision &Recall&Precision &Recall&Precision &Recall &Precision &Recall&Precision &Recall\\
 \hline
   AWS1&1& \B 0.5& \B 1&0.052& 1&0.041&1&0.045&1&0.047&1&0.125&1 &NA&NA&0.0087&1&0.009&1\\
 \hline
   AWS2&2   &0.13&1& \B 0.22& \B 1&  0.0042&1&0.1&0.5&0.18&1&0.11&1 &NA&NA&0.0062&1&0.04&1\\
 \hline
 AWS3&1& \B 1& \B 1&  0.37&  1&0.0181&1&0.0344&1&0.055&1&0&0 &NA&NA&0.005&1&0.006&1\\
 \hline
 Ambient temperature&1& 0.03&1& \B 0.0769&  \B 1&0.02&1&0.055&1&0&0&0&0 &NA&NA&0.01&1&0.02&1\\
 \hline
 GE1&3& 0.019&1&  0.048&  1&0.0357&1& \B 0.093& \B 1&0.041&0.33&0&0 &0.12&1&0.004&1&0.2&1\\
 \hline
 GE2&8& \B 1& \B 1&  0.66&  1& \B 1& \B 1& \B 1& \B 1&0&0&0.8&1 &0.8&1&0.16&1&0.034&1\\
 \hline
 Yahoo1&2& \B 0.076& \B 1&  0.0363&  1&0.0465&1&1&0.5&0.066&1&0.07&0.5 &0&0&0.005&1&0.009&1\\
 \hline
 Yahoo2&8& 0.75&0.375&  \B 0.8&  \B 1&1&0.375&1&0.25&0.19&0.625&0.10&0.25 &0&0&0.04&0.875&0.055&1\\
 \hline
 Yahoo3&8& \B 0.615& \B 1&  0.114&  0.675&0.088&1&0.023&0.25&0.11&0.875&0.15&0.62 &0.39&0.5&0.04&0.875&0.032&0.875\\
 \hline
 Yahoo5&9& 0.048&0.33&  0.1&  0.33&0.022&0.66&0.05&0.33&0&0&0.23&0.33 &\B 0.67& \B 1&0.029&0.66&0.029&0.66\\
 \hline
  Yahoo6&4& 0.12&1&  0.222&  1&0.0275&1&0.048&1&0&0&0.041&1 &\B 1& \B 1&0.0073&1&0.0075&1\\
 \hline
  Yahoo7&11& 0.096&0.54& \B  0.16&  \B 0.63&0.066&0.54&0.083&0.45&0.035&0.54&0.058&0.09 &0.33&0.29&0.0082&0.33&0.017&0.54\\
 \hline
Yahoo8&10&0.053&	0.7& \B	0.142& \B	0.8&	0.028&	0.3&0&0&0&0&0&0 &0.063 & 0.11 &0.01&0.6&0.010&0.6\\
\hline
 Yahoo9&8&1&	0.75&	\B 0.333& \B	1&	0.0208&	0.75&1&0.37&0&0&0.5&0.375 &0.07&0.8&0.04&1&0.047&1\\
 \hline
\end{tabular}
}
\caption{Performance comparison of various quantile LSTM techniques with other state of the art algorithms.}
\label{table:quantilecomp}
\end{table*}

\begin{table*}[!htb]
\renewcommand\thetable{4}
 \centering
 \resizebox{\textwidth}{!}{
 \begin{tabular}{|c|c|c|c|c|c|c|c|c|c|c|c|c|c|c|c|c|c|c|c|}
\hline
    Dataset&Anomaly&  \multicolumn{2}{c|}{iqr-LSTM}&\multicolumn{2}{c|}{Median-LSTM}&\multicolumn{2}{c|}{quantile-LSTM}&\multicolumn{2}{c|}{iForest} &\multicolumn{2}{c|}{Envelope} &\multicolumn{2}{c|}{Autoencoder} &\multicolumn{2}{c|}{GAN}&\multicolumn{2}{c|}{DAGMM} &\multicolumn{2}{c|}{DevNet}\\
   \hline
      &  & Precision &Recall& Precision &Recall&Precision &Recall&Precision &Recall&Precision &Recall&Precision &Recall&Precision &Recall&Precision &Recall &Precision &Recall\\
 \hline
   AWS\_syn1&11&0.769&	0.909&		0.687&	1&	1&	0.909&	0.034&	1&	0.10&	1&1&0.63& \B 0.84& \B 1&0.71&0.90 &0.09&0.73\\
 \hline
   AWS\_syn2&22   &0.7&	1&		\B 0.733& \B	1&		0.6875&	1&	0.065&	1&	0.33&	1&0.5&0.63&0.70&1&0.56&1 &0.44&0.27\\
 \hline
 AWS\_syn3&11& 1&	0.9	&	0.47&	1& \B	1&	\B 1&	0.025&	1&	0.072&	1&0.64&0.5&0.68&1&0&0 &0.2&0.45\\
 \hline
 
 GE\_syn1&13& 0.0093&	1&		0.203&	1&	0.071&	0.769&	0.0208&	1&		0.135&	1&0.23&0.11&0.25&0.61&0&0 & \B 0.33& \B 1\\
 \hline
 GE\_syn2&18& 0.0446&	1& \B	1&	\B 1& \B	1&	\B 1&	0.3&	1&		0.409	&1&1&0.38&0.9&1&0.9&1 &0.9&1\\
 \hline
 Yahoo\_syn1&12& \B 1& \B	1&		0.217&	0.833&	0.375&	1&		0.027&1&		0.056&	1&1&0.83& 0.31&1 &0.29&0.41 &0&0\\
 \hline
 Yahoo\_syn2&18& 0.181&	0.55&	0.653&	0.944&	1&	0.611&		\B 0.233& \B	1&	0.124&	1&1&0.42& 1&0.61& 0.55&0.61 &0&0\\
 \hline
 Yahoo\_syn3&18& 0.89&	0.94&		0.3333&	0.555& \B	0.6& \B	1&		0.0410&	1&		0.0762&	0.944&1&0.88& 0.81&0.71& 0.3&0.66 &0.17&0.63\\
 \hline
 Yahoo\_syn5&19& 0.081&	0.52&	0.521&	0.631&	0.0625&	0.578&		0.03125&	0.842&		0.0784	&0.842&0.15&0.47& 0.42&0.53& 0.52&0.52 & \B 0.73& \B 0.92\\
 \hline
  Yahoo\_syn6&14& 0.065&	0.85&		0.65&	0.928&	0.764&	0.928&	\B	0.01825& \B	1&	0.00761&	0.285&0.05&0.28& 0.8&0.29 &0.041&0.28 &0&0\\
 \hline
  Yahoo\_syn7&21& 0.61&	0.59&	0.375&	0.714&	0.411&	0.66&	\B	0.032&	\B 0.952&		0.052&	0.85&0.18&0.42& 0.14&0.38& 0.058&0.047 &0.11&0.64\\
 \hline
Yahoo\_syn8&20&0.32&	0.65&	\B	0.482&	\B 0.823&	0.197&	0.7&	0.0192&0.75&	0.023&	0.7&0.009&0.05& 0.25&0.1& 0&0 &0.23&0.64\\
\hline
 Yahoo\_syn9&18&1&	0.77&	\B	1&	\B	1&	1&	0.94&	0.053&	1&	0.048&	1&0.875&0.388& 0.72& 1& 0.57&0.22 &0.03&0.29\\
 \hline
\end{tabular}
}
\caption{Performance comparison of various quantile LSTM techniques on synthetic datasets with other state of the art algorithms.}
\label{table:quantilesyncomp}
\end{table*}

Table \ref{table:quantilesyncomp} shows the comparison with other baseline algorithms on multiple synthetic datasets. As in the previous table, Recall and Precision have been shown as performance metrics. As per these metrics, quantile-based approaches have outperformed \ac{if} and other deep learning based algorithms on 7 out of 13 datasets. If we consider the Precision alone, the quantile LSTM based techniques have demonstrated better performance on 10 synthetic datasets.  
There are multiple reasons for the better performance demonstrated by the quantile based LSTM approaches. First is the efficacy of the LSTM, which is well documented. Median-LSTM has detected the anomalies for each time period utilizing mean and standard deviation. It also has helped to capture the trend and seasonality. quantile-LSTM do not have any predefined threshold, which has improved their performance. Additionally, the flexibility of the parameter $\alpha$ in determining the shape of the activation helped in isolating the anomalies. This is evident from Fig \ref{fig:betaplot} which represents the variation in $\alpha$ values of the \ac{pef} function across the datasets.  $\alpha$ has been initialized to $1.5$ for all the datasets.
\subsection{Results-Non-Industrial Datasets}\label{appendix:nonindustrail}
We have tested our approach on non-industrial datasets shown in Table \ref{table:nonindustrial}. Here, Deviation Networks gives NA because it does not work for single anomaly containing datasets. 
On analysis of the results, we find that the quantile based technique is better in three of the seven datasets while Autoencoder is better for two of the seven datasets. 
\begin{table}[!htb]
\renewcommand\thetable{5}
 \centering
 \resizebox{\textwidth}{!}{
 \begin{tabular}{|c|c|c|c|c|c|c|c|c|c|c|c|c|c|}
\hline
    Dataset&Anomaly&  \multicolumn{2}{c|}{quantile-LSTM}&\multicolumn{2}{c|}{Autoencoder}&\multicolumn{2}{c|}{GAN} &\multicolumn{2}{c|}{DevNet}  &\multicolumn{2}{c|}{\ac{if}} &\multicolumn{2}{c|}{Envelope}  \\
   \hline
      &  & Precision &Recall& Precision &Recall&Precision &Recall&Precision &Recall &Precision &Recall&Precision &Recall\\
 \hline
 TravelTime$_{387}$ &3 & \B 0.011& \B 0.67 & 1&0.33 &0.024&0.33 & 0.01&0.33 & 0.0039&0.6667 &	0.0107&0.6667\\ 
 \hline
 TravelTime$_{451}$ &1 &0.006&1 & 0&0 & \B 0.016& \B 1 & NA&NA & 0.0028&1 & 0.0062&1\\ 
 \hline
 Occupancy$_{6005}$ &1 & \B 0.03& \B 1 & 0&0 &0.007&1 & NA&NA & 0.0019&1 & 0.0042&1\\ 
 \hline
 Occupancy$_{t4013}$ &2 & \B 0.06& \B 1 & 0.438&0.5 &0.014&0.5 & 0.02&1 & 0.0038&1 & 0.0078&1\\ 
 \hline
 Speed$_{6005}$ &1 &0.014&1 & \B 0.103& \B 1 &0.009&1 & NA&NA & 0.002&1 & 0.0038&1\\ 
 \hline
 Speed$_{7578}$ &4 &0.086&1 & \B 0.792& \B 1 &0.2&0.9 & 0.16&0.75 & 0.0153&1 & 0.0247&1\\ 
 \hline
 Speed$_{t4013}$ &2 &0.053&1 & 0.75&0.5 &0.043&1 & \B 0.1& \B 1 & 0.0036&1 & 0.007&1\\ 
 \hline
\end{tabular}
}
\caption{Performance comparison of quantile LSTM techniques on various non-industrial datasets}
\label{table:nonindustrial}
\end{table}

\subsection{Comparison between Elliot Function and \ac{pef}}\label{appendix:pefvsef}
In order to compare the performance of the Elliot function and parameterized Elliot function (PEF) as activation functions, we experimented with them by using them as activation functions in the LSTM layer of the models and comparing the results after they run on multiple datasets. The results are shown in Table \ref{table:pefvsef}.
\begin{table}[!htpb]
 \centering
 \renewcommand\thetable{6}
 \resizebox{0.8\textwidth}{!}{
 \begin{tabular}{|c|c|c|c|c|}
\hline
   Dataset &  \multicolumn{2}{c|}{Elliot Function}&\multicolumn{2}{c|}{Parameterized Elliot Function}\\
    \hline
    & Precision  & Recall & Precision  & Recall\\
    \hline
    AWS Dataset$_1$ & 0&0 & \B 0.041& \B1\\
    \hline
    AWS Dataset$_2$ & 0.002&1 & \B 0.0042& \B 1\\
    \hline
    AWS Dataset$_3$ & \B 0.04 & \B 1 & 0.0181&1\\
    \hline
    AWS DatasetSyn$_1$ & 0.02& 0.73 & \B 1& \B 0.909\\
    \hline
    AWS DatasetSyn$_2$ & 0.39&0.77 & \B 0.6875&\B 1\\
    \hline
    AWS DatasetSyn$_3$ & 0.06&0.73 & \B 1& \B 1\\
    \hline
    Yahoo Dataset$_1$ & 0.006&0.25 & \B 0.0465& \B 1\\
    \hline
    Yahoo Dataset$_2$ & \B  0.02& \B 1 & 1&0.375\\
    \hline
    Yahoo Dataset$_3$ & 0.05&1 & \B 0.088& \B 1\\
    \hline
    Yahoo Dataset$_5$ & 0.001&0.33 & \B 0.022& \B 0.66\\
    \hline
    Yahoo Dataset$_6$ & 0.002&0.17 & \B 0.0275&\B 1\\
    \hline
    Yahoo Dataset$_7$ & 0.03&0.09 & \B 0.066& \B 0.54\\
    \hline
    Yahoo Dataset$_8$ & \B 0.017& \B 0.4 & 0.028&0.3\\
    \hline
    Yahoo Dataset$_9$ & \B 0.43& \B 0.75 & 0.0208&0.75\\
    \hline
    Yahoo DatasetSyn$_1$ & 0.14&0.86 & \B 0.375&\B 1\\
    \hline
    Yahoo DatasetSyn$_2$ & \B 0.04& \B 0.72 & 1&0.611\\
    \hline
    Yahoo DatasetSyn$_3$ & 0.1&0.78 & \B 0.6& \B 1\\
    \hline
    Yahoo DatasetSyn$_5$ & 0.004&0.31 & \B 0.0625& \B 0.578\\
    \hline
    Yahoo DatasetSyn$_6$ & 0.015&0.69 & \B 0.764& \B 0.928\\
    \hline
    Yahoo DatasetSyn$_7$ & 0.35&0.43 & \B 0.411& \B 0.66\\
    \hline
    Yahoo DatasetSyn$_8$ & 0.024&0.5 & \B 0.197& \B 0.7\\
    \hline
    Yahoo DatasetSyn$_9$ & 0.27&0.67 & \B 1& \B0.94\\
    \hline
\end{tabular}
}
\caption{Comparison of Precision and Recall score for LSTM with Elliot Function and PEF as Activation Function}
\label{table:pefvsef}
\end{table}
According to the data gathered after running the models, we found that parameterized Elliot function has better Precision and Recall for as except for four of the datasets. Thus, we could conclude that using parameterized Elliot function as an activation function gave better performance for quantile-LSTM.

\subsection{Impact of Varying Thresholds}\label{sub:thresholds}
Deep-learning based algorithms such as Autoencoder~\cite{Sakurada:2019}, GAN~\cite{Zenati2018},
DAGMM~\cite{Zong2018} and DevNet~\cite{pang2019deep} consider upper threshold and lower thresholds on reconstruction errors or predicted value. To understand the impact of different thresholds on the performance, we have considered three baseline algorithms GAN, Autoencoder and Devnet. The baseline methods have considered three different sets of threshold values for upper and lower thresholds. The sets are shown in column head of tables \ref{table:gan_percentile}, \ref{table:autoenc_percentile} and \ref{table:devnet_percentile}, where the first threshold is the upper percentile and the second threshold is the lower percentile. In contrast, q-LSTM is robust against thresholds as data sets vary i.e. it captures all anomalies successfully within the $0.1$ and $0.9$ quantile threshold.

\begin{table}[!htpb]
 \centering
 \renewcommand\thetable{7}
 \resizebox{0.8\textwidth}{!}{
 \begin{tabular}{|c|c|c|c|c|c|c|}
\hline
   GAN &  \multicolumn{2}{c|}{99.25 and 0.75}&\multicolumn{2}{c|}{99.75 and 0.25}&\multicolumn{2}{c|}{99.9 and 0.1}\\
    \hline
    Dataset & Precision  & Recall & Precision  & Recall &Precision  & Recall\\
    \hline
    Yahoo Dataset$_1$ & 0.09&1 & 0.25&1 &0.5&1\\
    \hline
    Yahoo Dataset$_2$ & 0.348&1 & 0.333&0.375 &0.4&0.25\\
    \hline
    Yahoo Dataset$_3$ & 0.28&0.5 & 0.444&0.286 &0.28&0.5\\
    \hline
    Yahoo Dataset$_5$ & 0&0 & 0.375&0.333 &0.6&0.333\\
    \hline
    Yahoo Dataset$_6$ & 0.5&0.5 & 0.5&1 &0.182&1\\
    \hline
    Yahoo Dataset$_7$ & 0.154&0.364 & 0.3&0.273 &0.5&0.182\\
    \hline
    Yahoo Dataset$_8$ & 0.038&0.1 & 0.1&0.1 &0.25&0.1\\
    \hline
    Yahoo Dataset$_9$ & 0.192&0.625 & 0.5&0.625 &0.5&0.25\\
    \hline
\end{tabular}
}
\caption{Comparison of Precision and Recall score for GAN with varying thresholds for anomaly Upper Bound and Lower Bound}
\label{table:gan_percentile}
\end{table}

\begin{table}[!htpb]
 \centering
 \renewcommand\thetable{8}
 \resizebox{0.8\textwidth}{!}{
 \begin{tabular}{|c|c|c|c|c|c|c|}
\hline
   Autoencoders &  \multicolumn{2}{c|}{99.25 and 0.75}&\multicolumn{2}{c|}{99.75 and 0.25}&\multicolumn{2}{c|}{99.9 and 0.1}\\
    \hline
    Dataset & Precision  & Recall & Precision  & Recall &Precision  & Recall\\
    \hline
    Yahoo Dataset$_1$ & 0.5&0.07 & 0.5&0.036 &0.5&0.019\\
    \hline
    Yahoo Dataset$_2$ & 0.5&0.4 & 0.333&0.5 &0.2&0.5\\
    \hline
    Yahoo Dataset$_3$ & 0.44&0.5 & 0.4&0.5 &0.25&0.333\\
    \hline
    Yahoo Dataset$_5$ & 0.5&0.5 & 0.5&0.5 &0.5&0.5\\
    \hline
    Yahoo Dataset$_6$ & 0.5&1 & 1&1 &0.25&1\\
    \hline
    Yahoo Dataset$_7$ & 0.5&0.5 & 0.5&0.5 &0.5&0.5\\
    \hline
    Yahoo Dataset$_8$ & 0.875&0.875 & 0.375&0.375 &0.5&0.75\\
    \hline
    Yahoo Dataset$_9$ & 0.75&0.5 & 0.25&0.5 &0.5&0.5\\
    \hline
\end{tabular}
}
\caption{Comparison of Precision and Recall score for Autoencoders with varying thresholds for anomaly Upper Bound and Lower Bound}
\label{table:autoenc_percentile}
\end{table}

\begin{table}[!htpb]
 \centering
 \renewcommand\thetable{9}
 \resizebox{0.8\textwidth}{!}{
 \begin{tabular}{|c|c|c|c|c|c|c|}
\hline
   Devnet &  \multicolumn{2}{c|}{99.25 and 0.75}&\multicolumn{2}{c|}{99.75 and 0.25}&\multicolumn{2}{c|}{99.9 and 0.1}\\
    \hline
    Dataset & Precision  & Recall & Precision  & Recall &Precision  & Recall\\
    \hline
    Yahoo Dataset$_1$ & 0.002&1 & 0.002&1 &0.001&1\\
    \hline
    Yahoo Dataset$_2$ & 0.005&1 & 0.005&1 &0.005&1\\
    \hline
    Yahoo Dataset$_3$ & 0.0078&1 & 0.0078&1 &0.0078&1\\
    \hline
    Yahoo Dataset$_5$ & 0.111&0.5 & 0.333&0.5 &0.333&0.5\\
    \hline
    Yahoo Dataset$_6$ & 0.167&1 & 0.5&1 &0.5&0.667\\
    \hline
    Yahoo Dataset$_7$ & 0.054&0.2 & 0.125&0.2 &0.25&0.2\\
    \hline
    Yahoo Dataset$_8$ & 0&0 & 0&0 &0&0\\
    \hline
    Yahoo Dataset$_9$ & 0&0 & 0&0 &0&0\\
    \hline
\end{tabular}
}
\caption{Comparison of Precision and Recall score for Devnet with varying thresholds for anomaly Upper Bound and Lower Bound}
\label{table:devnet_percentile}
\end{table}

It is evident from the above tables that performance varies significantly based on the thresholds decided by the algorithm. Therefore it is very important to decide on a correct threshold that can identify all the probable anomalies from the dataset.
\subsection{Experiments on Normal Instances}\label{sub:Normal}
A relevant question to ask is: how would the anomaly detection methods perform on normal data instances that does not have any anomaly? 
We investigate this by removing anomalies from some data sets. We observe that on these data sets (AWS1, AWS2, AWS3, Yahoo1, Yahoo2, Yahoo3), q-LSTM and its variants reported very negligible false alarms (Average 40 false alarms) while other state-of-the-art methods, such as \ac{if}, Elliptic Envelope produce higher flag false positives. Elliptic envelope has reported, on average, 137 false alarms whereas \ac{if} reported an average of 209 false alarms across the datasets. Autoencoder and GAN, both have reported average false alarms 46 and 123 respectively, which is higher than the false positive rate of q-LSTM. This establishes the robustness of the proposed method.
\section{Related Work}\label{related}
Well-known supervised machine learning approaches such as Linear Support Vector Machines (SVM), Random Forest (RF), and Random Survival Forest (RSF)~\cite{Voronov, Verma} have been explored for fault diagnosis and the lifetime prediction of industrial systems. \cite{Anton} have explored SVM and RF to detect intrusion based on the anomaly in industrial data. 
Popular unsupervised approaches, such as  Anomaly Detection Forest \cite{Sternby2020AnomalyDF}, and K-means based Isolation Forest \cite{Karczmarek} try to isolate the anomalies from the normal dataset. These methods do not require labeled data. \cite{Karczmarek} considered K-means based anomaly isolation, but the approach is tightly coupled with a clustering algorithm. Anomaly Detection Forest like k-means based \ac{if}  requires a training phase with a subsample of the dataset under consideration. A wrong selection of the training subsample can cause too many false alarms.
The notion of ``likely invariants'' uses operational data to identify a set of invariants to characterize the normal behavior of a system, which is similar to our strategy. Such as an approach has been attempted to discover anomalies of cloud-based systems~\cite{Russo}. However, such an approach requires labeling of data and retuning of parameters when the nature of datasets vary.
Recently, \ac{dl} models based on auto-encoders, long-short term memory \cite{Erfani,RobustAutoencoder} are increasingly gaining attention for anomaly detection. \cite{Yin} have proposed an integrated model of \ac{cnn} and \ac{lstm} based auto-encoder for Yahoo Webscope time-series anomaly detection. For reasons unknown,~\cite{Yin} have taken only one Yahoo Webscope data to demonstrate their approach's efficacy.  The DeepAnT \cite{Munir} approach employs \ac{dl} methods and it uses unlabeled data for training. However, the approach is meant for time-series data sets such as Yahoo Webscope, Real traffic, AWS cloudwatch. A stacked \ac{lstm} \cite{Malhotra}  is used for time series anomaly prediction, and the network is trained on a normal dataset. The hierarchical Temporal Memory (HTM) method has been applied recently on sequential streamed data and compared with other time series forecasting models \cite{Osegi:2021}. The authors in \cite{Saurav:2018} have performed online time-series anomaly detection using deep RNN. The incremental retraining of the neural network allows to the adoption of concept drift across multiple datasets. There are various works~\cite{Forero:2019,Sperl2020}, which attempt to address the data imbalance issue of the anomaly datasets since anomalies are very rare and occur occasionally. Hence they propose semi-supervised approaches. However, the semi-supervised approach cannot avoid the expensive dataset labeling. Some approaches~\cite{zong2018deep} apply predefined thresholds, such as fixed percentile values to detect the anomalies. However, a fixed threshold value may not be equally effective on different domain datasets. Deep Autoencoding Gaussian Mixture Model (DAGMM) is an unsupervised DL-based anomaly detection algorithm~\cite{zong2018deep}, where it utilizes a deep autoencoder to generate a low-dimensional representation and reconstruction error for each input data point and is further fed into a Gaussian Mixture Model (GMM). 
Deviation Network(DevNet)~\cite{pang2019deep} is a novel method that harnesses anomaly scoring networks, Z-score based deviation loss, and Gaussian prior together to increase efficiency for anomaly detection. 
\section{Discussion and Conclusion }\label{sec:conclusion}
In this paper, we have proposed multiple versions of the SoTA anomaly detection algorithms along with a forecasting-based LSTM method. We have demonstrated that combining the quantile technique with LSTM can be successfully implemented to detect anomalies in industrial and non-industrial datasets without label availability for training. We have also exploited the parameterized Elliot activation function and shown anomaly distribution against quantile values, which helps in deciding the quantile anomaly threshold. The design of a flexible form activation, i.e., \ac{pef}, also helps in accommodating variance in the unseen data as the shape of the activation is learned from data. PEF, as seen in Table \ref{table:pefvsef} captures anomalies better than vanilla Elliot. The quantile thresholds are generic and will not differ for different datasets. The proposed techniques have addressed the data imbalance issue and expensive training dataset labeling in anomaly detection. These methods are useful where data is abundant. Traditional deep learning-based methods use classical conditional means and assume random normal distributions as the underlying structure of data. These assumptions make the methods vulnerable to capturing the uncertainty in prediction and make them incapable of modeling tail behaviors. Quantile in LSTM (for time series data) is a  robust alternative that we leveraged in isolating anomalies successfully. This is fortified by the characteristics of quantiles making very few distributional assumptions. 
The distribution-agnostic behavior of Quantiles turned out to be a useful tool in modeling tail behavior and detecting anomalies. Anomalous instances, by definition, are rare and could be as rare as just one anomaly in the entire data set. Our method detects such instances (singleton anomaly) while some, recent state of art algorithms such as DAGMM require at least two anomalies to be effective. Extensive experiments on multiple industrial timeseries datasets (Yahoo, AWS, GE, machine sensors, Numenta and VLDB Benchmark data) and non-time series data
show evidence of effectiveness and superior performance of LSTM-based quantile
techniques in identifying anomalies. The proposed methods have a few drawbacks \begin{enumerate*}
    \item quantile based \ac{lstm} techniques are applicable only on univariate datasets. 
    \item A few of the methods such as quantile-LSTM, iqr-LSTM have a dependency on multiple thresholds.
\end{enumerate*} 
We intend to introduce the notion of multiple dimensions in our quantile-based approaches to detect anomalies in multivariate time series data in the future.
\bibliography{master.bib}

\bibliographystyle{plain}

\newpage
\begin{appendices}
\section{Probability Bound}\label{appendix:pobabilitybound}
Table \ref{table:probabilitybound} shows the probabilities of the anomalies at various quantile thresholds. The table has demonstrated all the 32 datasets considered as part of the experiment section.
\begin{table}[!htb]
\renewcommand\thetable{10}
 \centering
 \resizebox{\textwidth}{!}{
 \begin{tabular}{|c|c|c|c|c|c|}
\hline
    Dataset&  $\mathcal{P(E}>0.95)$&$\mathcal{P(E}>0.9)$&$\mathcal{P(E}>0.75)$&$\mathcal{P(F}<0.25)$ &$\mathcal{P(F}<0.10)$  \\
   \hline
 AWS Dataset$_1$&0&0.01&0.004&0&0\\
 \hline
 AWS Dataset$_2$&0&0.1&0.1&0&0\\
 \hline
 AWS Dataset$_3$&0&0.007&0.0032&0&0\\
 \hline
 Yahoo Dataset$_1$&0&0.014&0.005&0&0\\
 \hline
 Yahoo Dataset$_2$&0&0.105&0.062&0&0\\
 \hline
 Yahoo Dataset$_3$&0&0.103&0.076&0&0\\
 \hline
 Yahoo Dataset$_4$&0&0.014&0.0055&0&0\\
 \hline
 Yahoo Dataset$_5$&0&0.043&0.016&0&0\\
 \hline
 Yahoo Dataset$_6$&0&0.028&0.011&0&0\\
 \hline
 Yahoo Dataset$_7$&0&0.047&0.018&0.0069&0.017\\
 \hline
 Yahoo Dataset$_8$&0&0.011&0.004&0.016&0.041\\
 \hline
 Yahoo Dataset$_9$&0&0.017&0.0069&0.011&0.029\\
 \hline
  Sensor Dataset$_1$&0&0.0344&0.0135&0&0\\
 \hline
 Sensor Dataset$_2$&0&0&0&0.013&0.033\\
 \hline
 GE Dataset$_1$&0&0.003&0.002&0&0\\
 \hline
 GE Dataset$_2$&0&0.05&0.042&0&0\\
 \hline

 AWS Datasetsyn$_1$&0&0.08&0.035&0&0\\
 \hline
 AWS Datasetsyn$_2$&0&0.08&0.035&0&0\\
 \hline
 AWS Datasetsyn$_3$&0&0.1&0.1&0&0\\
 \hline
 Yahoo Datasetsyn$_1$&0& 0.074&	0.034&		0&	0\\
 \hline
 Yahoo Datasetsyn$_2$&0& 0.21&	0.15&	0&	0\\
 \hline
 Yahoo Datasetsyn$_3$& 0&	0.13&		0.11&	0&	0\\
 \hline
 Yahoo Datasetsyn$_5$&0& 0.08&	0.036&	0&	0\\
 \hline
  Yahoo Datasetsyn$_6$&0& 0&	0&		0.025&	0.062\\
 \hline
  Yahoo Datasetsyn$_7$&0& 0.047&	0.018&	0.03&	0.076\\
 \hline
Yahoo Datasetsyn$_8$&0&0.034&	0.015&		0.026&	0.051\\
\hline
 Yahoo Datasetsyn$_9$&0&	0.017&		0.0069&		0.034&	0.088\\
 \hline
 Sensor Datasetsyn$_1$&0&0&0&0.108&0.39\\
 \hline
 Sensor Datasetsyn$_2$&0&0&0&0.146&0.36\\
 \hline
 GE DatasetSyn$_1$&0&0.017&0.0104&0&0\\
 \hline
 GE DatasetSyn$_2$&0&0.11&0.096&0&0\\
 \hline

\end{tabular}
}
\caption{Various probability values on different quantile threshold parameters.}
\label{table:probabilitybound}
\end{table}
\section{Lemma}\label{appendix:lemmaproof}
\textbf{Lemma 1:}\\
For an univariate dataset $\mathcal{D}$, 
the probability of an anomaly $\mathcal{P(A)}=\mathcal{P}(\mathcal{E} > \alpha_{high}) +\mathcal{P(F}<\alpha_{low})$, where $\alpha_{high}, \alpha_{low}$ are the higher and lower level quantile thresholds respectively.

\begin{proof}
A datapoint is declared an anomaly if its quantile value is higher than $\alpha_{high}$ or lower than $\alpha_{low}$. Here $\alpha_{high}$, $\alpha_{low}$ are the higher and lower quantile threshold value.
$\mathcal{P}(\mathcal{E} > \alpha_{high})$ is the probability of an anomaly whose quantile value is higher than $\alpha_{high}$. On the other side, $\mathcal{P(F}<\alpha_{low})$ is the probability of quantile value of anomalous datapoint lower than $\alpha_{low}$. Hence the presence of an anomaly in a dataset is possible if one of the events is true. Therefore
\begin{align*}
    &\mathcal{P(A)}=P(\mathcal{E} > \alpha_{high} \cup \mathcal{F}<\alpha_{low})\\
    &\mathcal{P(A)}=\mathcal{P}(\mathcal{E} > \alpha_{high})+\mathcal{P(F}<\alpha_{low})-P(\mathcal{E} > \alpha_{high} \cap \mathcal{F}<\alpha_{low})
\end{align*}
Both the events $\mathcal{E, F}$are mutually exclusive. Hence the above Equation can be written as
\begin{equation}\label{eq:quantilethreshold}
     \mathcal{P(A)}=\mathcal{P}(\mathcal{E} > \alpha_{high})+\mathcal{P(F}<\alpha_{low})
\end{equation}

\end{proof}
\section{Backpropagation of LSTM with PEF}\label{appendix:backpropa}
\begin{figure}[!tbp]
  \centering
    \includegraphics[width=0.5\textwidth]{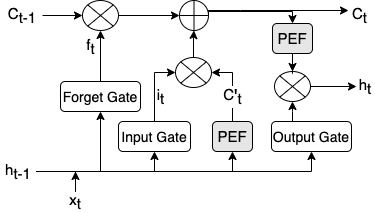}
\caption{LSTM cell structure }
\label{fig:lstm}
\end{figure}
For backward propagation, it is required  to compute the derivatives for all major components of the LSTM. $J$ is the cost function and the relationship between $v_t$ and hidden state $h_t$ is $v_t=w_v*h_t+b_v$. The predicted value $y'=softmax(v_t)$. The derivative of the hidden state can be shown as follow
\begin{align*}
    &\frac{\partial J}{\partial h_t}=\frac{\partial J}{\partial v_t}\frac{\partial v_t}{\partial h_t}\\
    &\frac{\partial J}{\partial h_t}=\frac{\partial J}{\partial v_t}\frac{\partial (w_v*h_t+b_v)}{\partial h_t}\\
    &\frac{\partial J}{\partial h_t}=\frac{\partial J}{\partial v_t}w_v
\end{align*}
The variable involved in the output gate is $o_t$.
\begin{align*}
    &\frac{\partial J}{\partial o_t}=\frac{\partial J}{\partial h_t}\frac{\partial h_t}{\partial o_t}\\
    &\frac{\partial J}{\partial o_t}=\frac{\partial J}{\partial h_t}\frac{\partial (o_t*PEF(C_t))}{\partial o_t}\\
    &\frac{\partial J}{\partial o_t}=\frac{\partial J}{\partial h_t}PEF(C_t)
\end{align*}
$C_t$ is the cell state and the chain rule for cell state can be written as 
\begin{align}
    &\frac{\partial J}{\partial C_t}=\frac{\partial J}{\partial h_t}\frac{\partial h_t}{\partial C_t}
\end{align}
$\frac{\partial J}{\partial h_t}$ value already we have calculated as part of hidden state equation.
\begin{align*}
    &\frac{\partial h_t}{\partial C_t}=\frac{\partial (o_t*PEF(C_t))}{\partial C_t}\\
    &=\frac{\alpha o_t}{(|C_t|+1)^2}
\end{align*}
After setting the value of $\frac{\partial h_t}{\partial C_t}$ in equation 4
\begin{align}
    \frac{\partial J}{\partial C_t}=\frac{\partial J}{\partial h_t}\frac{\alpha o_t}{(|C_t|+1)^2}
\end{align}
 The chain rule for $c\hat{}_t$ is 
\begin{align*}
    &\frac{\partial J}{\partial c\hat{}_t}=\frac{\partial J}{\partial C_t}\frac{\partial C_t}{\partial c\hat{}_t}
\end{align*}
We need to derive only $\frac{\partial C_t}{\partial c\hat{}_t}$ since $\frac{\partial J}{\partial C_t}$ is available from equation 5.
\begin{align*}
   & \frac{\partial C_t}{\partial c\hat{}_t}=\frac{\partial (f_t*C_{t-1}+c\hat{}_t*i_t)}{\partial c\hat{}_t}\\
   &=i_t
\end{align*}
After replacing the value of $\frac{\partial C_t}{\partial c\hat{}_t}$
\begin{align*}
    \frac{\partial J}{\partial c\hat{}_t}=\frac{\partial J}{\partial C_t}*i_t
\end{align*}
Similar way $ \frac{\partial J}{\partial a_c}=\frac{\partial J}{\partial c\hat{}_t}*\frac{\alpha}{(|a_c|+1)^2}$
The following derivatives for input gate
\begin{align*}
    &\frac{\partial J}{\partial i_t}=\frac{\partial J}{\partial C_t}c\hat{}_t\\
    &\frac{\partial J}{\partial a_i}=\frac{\partial J}{\partial C_t}c\hat{}_t(i_t(1-i_t))
\end{align*}
For forget gate, below are the derivatives
\begin{align*}
    &\frac{\partial J}{\partial f_t}=\frac{\partial J}{\partial C_t}C_{t-1}\\
    &\frac{\partial J}{\partial a_f}=\frac{\partial J}{\partial C_t}C_{t-1}(f_t(1-f_t))
\end{align*}
$Z_t$ is the concatenation of the $h_{t-1},x_t$. The derivatives of the weights are as follow
\begin{align*}
     &\frac{\partial J}{\partial w_f}=\frac{\partial J}{\partial a_f}Z_t\\
      &\frac{\partial J}{\partial w_i}=\frac{\partial J}{\partial a_i}Z_t\\
        &\frac{\partial J}{\partial w_v}=\frac{\partial J}{\partial v_t}h_t\\
          &\frac{\partial J}{\partial w_o}=\frac{\partial J}{\partial a_o}Z_t\\
\end{align*}
\subsection{Parameterized Elliot Function}
One of the major benefit of the parameterized Elliot function is that it further decreases the rate of saturation in comparison to the non-parameterize Elliot function. We have applied one parameter $\alpha$, which controls the shape of the Elliot function. There will be different derivatives if we apply parameterize Elliot function in LSTM. 
\begin{align*}
    PEF=\frac{\alpha x}{1+|x|}
\end{align*}
After the introduction of the PEF, the hidden state equation is as follow
\begin{align*}
    h_t=O_t\alpha_c PEF(C_t)
\end{align*}
As per the chain rule, the derivative for $\alpha_c$ will be
\begin{align*}
   & \frac{\partial J}{\partial \alpha_c}=\frac{\partial J}{\partial h_t}\frac{\partial O_t\alpha_c Elliot(C_t)}{\partial \alpha_c}\\
    & \frac{\partial J}{\partial \alpha_c}=\frac{\partial J}{\partial h_t}O_t *Elliot(C_t)
\end{align*}
After each iteration, the $\alpha_c$ is updated as per equation \ref{eq:hiddenalpha}.

\begin{equation}\label{eq:hiddenalpha}
    \alpha_c^{(n+1)}=\alpha_c^n+\delta*\frac{\partial J}{\partial \alpha_c}
\end{equation}
Similarly, we can derive  $\alpha_c\hat{}$ and update the parameter.

\section{Intuition with An Example:}\label{appendix:example} 

It is hypothesized that, in neural networks, the logistic layer output softmax(b+Wh) might initially rely more on the biases b and hence push the activation value h towards 0, thus resulting in error gradients of smaller values. They referred to this as the saturation property of neural networks. This results in slower training and prevents the gradients from propagating backward until the layers close to the input learns. This saturation property is observed in the sigmoid. The sigmoid is non-symmetric around zero and obtains smaller error gradients when the sigmoid outputs a value close to 0. Similarly, tanh in all layers tends to saturate towards 1, which leads to layer saturation. All the layers attain a particular value, which is detrimental to the propagation of gradients.
However, this issue of attaining saturation would be less pronounced in cases where two different activation functions are used. Since each activation function behaves differently in terms of gradients, i.e., sigmoid outputs are in the range [0,1], and the gradients are minimum at the maximum and minimum values of the function. The $\tanh$ on the other hand, has minimum gradients at -1 and 1 and reaches its maximum at 0. Therefore, even if the layers begin to saturate to a common value, some of the layers would escape the saturation regime of their activations and would still be able to learn essential features. As an outcome, this might result in fewer instances of vanishing gradients. This assumption would mean that networks with two different activations would learn faster and converge faster to a minima, and the same premise is supported by a Convergence study (details in section V). As demonstrated by Glorot and Bengio, if the saturation ratio of layers is less pronounced, it leads to better results in terms of accuracy. A standard neural network with N layers is given by $ h^{l} = \sigma(h^{l-1}W^{l}+b)$ and $s^{l} = h^{l-1}W^{l}+b$. Here $h^{l}$ is the output of the first hidden layer, $\sigma$ is a non-linear activation function, and b is the bias. We compute the gradients as $\frac{\partial Cost}{\partial s^{l}_{k}} = f'(s^{l}_{k}) W^{l}_{k,\cdot}\frac{\partial Cost}{\partial s^{l+1}}$; $ \frac{\partial Cost}{\partial W^{l}_{m,n}} = z^{i}_{l}\frac{\partial Cost}{\partial s^{l}_{k}}$. Now, we find the variances of these values. As the network propagates, we must ensure that the variances are equal to keep the information flowing. Essentially, when $\forall(l,l'), Var[h^l] = Var[h^{l^{'}}]$, it ensures that forward propagation does not saturate, and when $\forall(l,l'), Var[\frac{\partial Cost}{\partial s^{l}}] = Var[\frac{\partial Cost}{\partial s^{l^{'}}}]$, it ensures that backward propagation flows at a constant rate. Now, what remains is to calculate these variance values. Let us consider an elaborate example. 

Firstly, we attempt to find variance for two sigmoid activations in a network. The derivative of each activation output is approximately 0.25($\sigma'(0) = 0.25$), as the weights are uniformly initialized, and the input features are assumed to have the same variance. Hence,
\begin{equation*}
    f'(s^{l}_{k}) = 0.25
\end{equation*}
\begin{equation*}
    Var[z^{2}] = Var[x]((0.25)^{2}n_{1}Var[W^{1'}]*(0.25)^{2}n_{2}Var[W^{2'}])
\end{equation*}
We see that this diminishing factor of $0.25^{N}$ steeply drops the variance during the forward pass.  Similarly, we observe that the gradient, 
\begin{equation*}
    \frac{\partial Cost}{\partial s^{l}_{k}} = f'(s^{l}_{k}) W^{l}_{k,\cdot}\frac{\partial Cost}{\partial s^{l+1}}
\end{equation*}
has $f'(s^{l}_{k})$ as one of the factors, and thus the diminishing factor is tied to the variance. Even when $N=2$ the variance reduces by a factor of $4^{4} = 256$. \\
Let's compute variance for neural network with two hidden layers using sigmoid and tanh activations. For tanh, if the initial values are uniformly distributed around 0, the derivative is $f'(s^{l}_{k}) = 1$. Therefore, the variance for the second layer output is, 
$Var[z^{2}] = Var[x]*((0.25)^{2}*n_{1}*Var[W^{1'}]*n_{2}*Var[W^{2'}])$.
We see that the diminishing factor is just $4^{2} = 16$, and this results in a much better variance when compared to the previous case. Therefore, using different AFs instead of the same implies a reduction in vanishing gradients and results in a much better flow of information because the variance value is preserved for longer.
\newpage
\section{Dataset Properties}\label{appendix:datasetchar}
\begin{table}[!htb] 
\renewcommand\thetable{11}
\centering
\resizebox{\textwidth}{!}{
\begin{tabular}{|l|l|l|l|l|l|l|l|}
\hline
Dataset Name    & Anomaly\% & Size & Missing Data & Modal    & Distribution & \#Variables & TW/ Period\\\hline
\multicolumn{8}{|c|}{Publicly available actual industrial data}                                      \\\hline
AWS Dataset$_1$ & 0.09       & 1049         & No           & Unimodal & Weibull      & Univariate&84/168  \\\hline
AWS Dataset$_2$ & 0.08 &2486&No&Unimodal&Weibull&Univariate&38/152 \\\hline
AWS Dataset$_3$ & 0.066&1499&No&Multimodal& Weibull&Univariate&38/152 \\\hline
Yahoo Dataset$_1$ & 0.14&1421&No&Unimodal &Weibull&Univariate&20/60 \\\hline
Yahoo Dataset$_2$  & 0.54&1462&No&Unimodal&Gamma&Univariate&30/90 \\\hline
Yahoo Dataset$_3$ & 0.55 &1440&No&Unimodal&Weibull&Univariate&10/120\\\hline
Yahoo Dataset$_4$ & 0.28&1422&No&Bimodal&Weibull&Univariate&105/210 \\\hline
Yahoo Dataset$_5$ & 0.63&1421&No&Bimodal&Log-normal&Univariate&24/72 \\\hline
Yahoo Dataset$_6$ & 0.28&1421&No&Multimodal&Weibull&Univariate&74/148\\\hline
Yahoo Dataset$_7$ & 0.53&1680&No&Unimodal&Weibull&Univariate&125/250 \\\hline
Yahoo Dataset$_8$ & 0.59&1680&No&Unimodal&Log-normal&Univariate&116/232\\\hline
Yahoo Dataset$_9$ & 0.47&1680&No&Unimodal&Weibull&Univariate&30/90 \\\hline
Machine Temperature Dataset & 0.19&501&No&Multimodal&Weibull&Univariate&38/114 \\\hline

\multicolumn{8}{|c|}{Private actual industrial data}                                      \\\hline
GE Dataset$_1$ &  0.18&1609&No&Unimodal&Exponential&Univariate&117/234\\ \hline
GE Dataset$_2$ & 1.47&544&No&Multimodal&Weibull&Univariate  &50/150   \\\hline
\multicolumn{8}{|c|}{Publicly available synthetic industrial data}                            \\\hline 
AWS DatasetSyn$_1$  & 1.03 &1059&No&Unimodal&Exponential&Univariate&84/168 \\\hline
AWS DatasetSyn$_2$ & 0.877&2506 &No&Unimodal&Weibull&Univariate&38/152 \\\hline
AWS DatasetSyn$_3$ & 0.72 &1509&No&Unimodal&Gamma&Univariate&38/152 \\\hline
Yahoo DatasetSyn$_1$ & 0.83 &1431 &No&Unimodal&Weibull&Univariate&20/60 \\\hline
Yahoo DatasetSyn$_2$  & 1.22&1472&No&Unimodal&Gamma&Univariate&30/90 \\\hline
Yahoo DatasetSyn$_3$ & 1.24&1450&No&Unimodal&Weibull&Univariate&10/120\\\hline
Yahoo DatasetSyn$_4$ & 0.977&1432&No&Bimodal&Weibull&Univariate&105/210\\\hline
Yahoo DatasetSyn$_5$ & 1.32&1431&No&Bimodal&Weibull&Univariate&24/72 \\\hline
Yahoo DatasetSyn$_6$ & 0.97&1431&No&Multimodal&Weibull&Univariate&74/148\\\hline
Yahoo DatasetSyn$_7$ & 1.12&1690&No&Unimodal&Weibull&Univariate&125/250 \\\hline
Yahoo DatasetSyn$_8$ & 1.18&1690&No&Multimodal&Log-normal&Univariate&116/232\\\hline
Yahoo DatasetSyn$_9$ & 1.065&1690&No&Bimodal&Weibull&Univariate&30/90\\\hline
\multicolumn{8}{|c|}{Private synthetic industrial data}                                      \\\hline
GE DatasetSyn$_1$ &  0.80&1619&No&Unimodal&Log-normal&Univariate&117/232 \\\hline
GE DatasetSyn$_2$ & 3.30&554&No&Multimodal&Exponential&Univariate&50/150 \\\hline

\end{tabular}
}
\caption{Anomaly Dataset Properties.}
\label{table:datasetschar}
\end{table}

\section{VLDB}\label{appendix:vldb}

Using the VLDB Benchmark, we generated a timeseries dataset of 5500 datapoints containing 40 anomalies. We used various Deep Learining based algorithms on this generated non-industrial dataset. From table \ref{table:vldb}, we can clearly see that in terms of recall our proposed algorithm(Median-LSTM) works almost as well as GAN and far better than the other two. While in terms of precision Median-LSTM gives the best values, thus performing better than the other algorithms overall.

\begin{table}[!htpb]
 \centering
 \renewcommand\thetable{12}
 \resizebox{0.5\textwidth}{!}{
 \begin{tabular}{|c|c|c|}
\hline
   VLDB Dataset &  Precision & Recall\\
   \hline
    Median-LSTM & 0.513&0.976\\
    \hline
    GAN & 0.0072&1 \\
    \hline
    Autoencoders & 0.013&0.025 \\
    \hline
    Devnet & 0.0357&0.158\\ 
    \hline
\end{tabular}

}
\caption{Comparison of Precision and Recall score for VLDB generated dataset for various deep learning anomaly detection techniques.}
\label{table:vldb}
\end{table}
%

%


\end{appendices}
\end{document}